\documentclass[runningheads]{llncs}
\usepackage{graphicx}
\usepackage{float}
\usepackage{listings}
\usepackage{xcolor}
\usepackage{amsmath}
\usepackage{bbm}
\usepackage{multirow}
\usepackage{dsfont}
\usepackage{lipsum}
\usepackage[ruled,noend]{algorithm2e}

\begin{document}
\title{RR-CP: Reliable-Region-Based Conformal Prediction for Trustworthy Medical Image Classification}
%
%
%
%
%
\author{Yizhe Zhang\inst{1}, Shuo Wang\inst{2,3}, Yejia Zhang\inst{4}, Danny Z. Chen\inst{4}}
\institute{School of Computer
Science and Engineering, Nanjing University of Science and Technology, Nanjing, Jiangsu 210094, China \and Digital Medical Research Center, School of Basic Medical Sciences, Fudan University, Shanghai 200032, China \and Shanghai Key Laboratory of MICCAI, Shanghai 200032, China \and Department of Computer Science and Engineering, University of Notre Dame, Notre Dame, IN 46556, USA \\\email{yizhe.zhang.cs@gmail.com; shuowang@fudan.edu.cn;\\ yzhang46@nd.edu; dchen@nd.edu}}
\titlerunning{Reliable-Region-Based Conformal Prediction}
\maketitle              
\begin{abstract}
Conformal prediction (CP) generates a set of predictions for a given test sample such that the prediction set almost always contains the true label (e.g., 99.5\% of the time). CP provides comprehensive predictions on possible labels of a given test sample, and the size of the set indicates how certain the predictions are (e.g., a set larger than one is `uncertain'). Such distinct properties of CP enable effective collaborations between human experts and medical AI models, allowing efficient intervention and quality check in clinical decision-making. In this paper, we propose a new method called Reliable-Region-Based Conformal Prediction (RR-CP), which aims to impose a stronger statistical guarantee so that an extremely low error rate (e.g., 0.5\%) can be achieved in the test time, and under this constraint, the size of the prediction set is optimized to be small. We consider a small prediction set size an important measure only when the low error rate is achieved. Experiments on five public datasets show that our RR-CP performs well: with a reasonably small-sized prediction set, it achieves the user-specified low error rate significantly more frequently than exiting CP methods.

\end{abstract}
\section{Introduction}
\begin{figure}[t]
\centering
\includegraphics[width=0.76\textwidth]{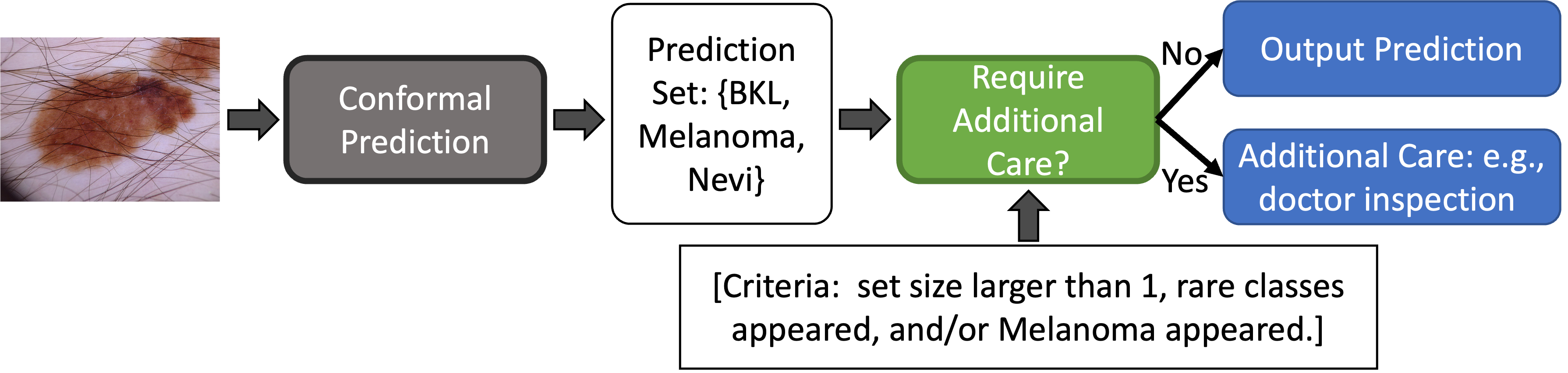}
\caption{Illustrating a practical CP-based workflow for trustworthy medical image classification: Conformal prediction works together with medical doctors.} 
\label{fig:overview}
\end{figure}

\vspace{-0.1cm}
Deep learning (DL) models nowadays have become go-to approaches for medical image classification in computer-aided diagnosis. Despite their successes in many medical AI benchmarks, the reliability and trustworthiness of DL models are still one of the major concerns and obstacles in applying DL-based methodologies in clinical practice. Recently, conformal prediction (CP) has drawn lots of attention for improving the trustworthiness of medical AI~\cite{lu2022improving,lu2022fair,angelopoulos2021gentle}. CP generates not only a single prediction label but a set of possible labels (containing one or multiple labels) for a given test image sample, and the true label is guaranteed (in the statistical sense) to be within the prediction set. CP enables effective collaborations between human experts and AI models~\cite{babbar2022utility}, allowing efficient intervention and quality check in clinical decision-making (see Fig.~\ref{fig:overview}).

More formally, given an image sample $x$ to a classification model $f$, we first obtain a probability output vector $\pi = f(x)$. A CP method then performs on top of this output to generate a prediction set of possible labels to which the sample $x$'s true label $y$ may belong: $PS(x)=CP(f(x))$. Traditionally, a CP method is designed to achieve the following goal: 
\begin{equation}
1- \alpha \leq P (Y_{test} \in PS(X_{test})) \leq 1 - \alpha + \epsilon,
\label{eq:1}
\end{equation}
where $Y_{test}$ is the true labels of test samples $X_{test}$, $\alpha$ is a user-specified error rate and should be set to a small value 
for critical applications (e.g., medical image classification), and $\epsilon>0$ is a small value for making the coverage tight.

A trivial solution for constructing a CP model is to let it produce all the possible class labels for any given sample with a $1-\alpha$ probability. Clearly, this would give little useful information on any test sample 
as the model almost always generates the same full prediction set for a test sample. To avoid such a trivial solution, the size of the prediction set must be considered when designing a CP model. A general goal that includes the prediction set size is to satisfy the constraints in Eq.~(\ref{eq:1}) using a prediction set of the smallest possible size. 

Suppose a calibration set $D_{cali} = (X_{cali}, Y_{cali})$, with $X_{cali} = \{x_1, x_2, \dots, x_n\}$ and $Y_{cali}=\{y_1, y_2, \dots, y_n\}$, is given for a $K$-class classification problem ($y_i\in \{1,2, \dots, K\}$). For a classifier $f$ (e.g., a ResNet classifier), a classic CP method~\cite{vovk1999machine} computes a score for each sample $x_i$ using $s(x_i,y_i) = 1-\pi_i[y_i]$, where $\pi_i=f(x_i)$ and $\pi_i[y_i]$ is the predicted probability of the ground truth $y_i$. The method then collects all the scores for the samples in $D_{cali}$ and finds the $1-\alpha$ quantile of the collected scores, denoted as $q_{\alpha}$. For a new sample $x_{test}$, with the classifier $f$ applied on top of it, the prediction set is generated by selecting any class label $k$ ($k \in \{1,2,\dots, K\}$) with the score $\pi_{test}[{k}] \geq 1 - q_{\alpha}$. In~\cite{vovk1999machine}, it was shown that using the above procedure, when the samples in a test set $D_{test} = (X_{test}, Y_{test})$ and the samples in the set $D_{cali}$ are independent and identically distributed (i.i.d.), then $
1- \alpha \leq P (Y_{test} \in PS(X_{test}))$ holds. Although this is a great theoretical result, two key issues remain: (1) The i.i.d. assumption is often too strong, and in practice, test data may not be sampled from the exactly same distribution of the data used to find $q_{\alpha}$. (2) No explicit control on the size of the prediction set: The size of the prediction set yielded in the above procedure depends entirely on the quality of the score function, which depends on the quality of the class probability output from the classifier $f$. If the probability output is entirely random, then the size of the prediction set can be arbitrarily large (up to $K$). 

Recently, Adaptive Prediction Set (APS)~\cite{romano2020classification,angelopoulosuncertainty} modified the classic CP method by accumulating class probabilities when computing scores. More formally, for each sample $(x_i,y_i)$ in $D_{cali}$, APS first sorts the class probabilities in each $\pi_i$ (from high to low) and obtains $ \tau_i = argsort ([\pi_i[1], \pi_i[2], \dots, \pi_i[K]])$, where $\tau_i$ is a permutation of $\{1, ..., K\}$ resulted from the sorting. The score for a sample $(x_i, y_i)$ is computed as $s(x_i, y_i) = \sum_{k=1}^{k^*}\pi_i[\tau(k)]$, with $\tau_i(k^*)=y_i$. With this new score function, the remaining procedure for computing $q_{\alpha}$ and using the computed $q_{\alpha}$ for getting the prediction set on a new sample in the test set $D_{test}$ is largely the same as the above classic CP method. APS aims to improve the robustness of CP by improving the score function and making a fuller utilization of the probability output. For multi-label classification tasks where a sample can be associated with more than one true label, a method \cite{fisch2022conformal} proposed to reduce false positives in the prediction set by trading with coverage, and demonstrated practical usage of this design in screening molecule structures. Recently, the CP formula was generalized using a more general term/formula (``risk'') to represent terms such as the error rate in classification~\cite{angelopoulos2022conformal}.

In medical image analysis, 
a most critical consideration of CP is to ensure that the actual error rate in deployment is no bigger than the user-specified one when constructing a CP method. If the user-specified error rate is not attained in test time, a smaller-sized prediction set is meaningless when the true label is not in the prediction set. Hence, we propose a new method called Reliable-Region-based Conformal Prediction (RR-CP), which puts satisfying the user-specified error rate as the first priority and reducing the prediction set size as the second one. We estimate a set of reliable regions (related to ~\cite{zhang2022usable}) for set predictions to ensure when a prediction set falls into a particular reliable region, this prediction set is guaranteed (statistically) to contain the true label. Overall, RR-CP can be viewed as a constrained optimization approach in which the constraint is to satisfy the user-specified error rate and the objective is to minimize the set size.

This paper contributes to trustworthy medical AI in the following ways. (1) We identify a crucial aspect of applying conformal prediction in medical AI, which is the expectation that a CP method should achieve an extremely low error rate (e.g., 0.5\%) during deployment. Such an extremely low error rate had not been addressed by previous CP methods. (2) In order to fulfill the low error rate requirement, we develop a novel method called reliable-region-based conformal prediction (RR-CP) which by design exerts a stronger correctness guarantee for achieving the desired minimal error rate. (3) Theoretical analyses demonstrate the distinct value of our proposed RR-CP, and experimental results show that RR-CP satisfies the desired low error rates significantly more often than known CP methods.

\section{Reliable-Region-Based Conformal Prediction}\label{sec:method}
In Section~\ref{sec:method-p1}, we first give a high-level elucidation of the overall goal, principles, and ideas of our proposed RR-CP. We then give detailed descriptions of the process and algorithm in Section~\ref{sec:method-p2}. In Section~\ref{sec:method-p3}, we provide further theoretical analyses and justifications for our RR-CP method.
\vspace{-0.3cm}
\subsection{Principles and Overall Ideas}\label{sec:method-p1}
We aim to produce a prediction set for test samples to attain the following goal: 
\begin{equation}
1- \alpha \leq P ( Y_{test} \in PS(X_{test})).
\label{eq:2}
\end{equation}
Note that Eq.~(\ref{eq:2}) is different from the original goal in Eq.~(\ref{eq:1}) as we drop the constraint of $P( Y_{test} \in PS(X_{test})) \leq 1-\alpha + \epsilon$, where $\epsilon$ is a small term. In~\cite{vovk1999machine}, it was shown that the classic CP method satisfies this constraint with $\epsilon = \frac{1}{n+1}$, where $n$ is the number of samples used in estimating the parameters of the CP method. Usually in critical applications such as medical image classification, the value of $\alpha$ is expected to be small (e.g., 0.005). Suppose we set $\alpha = 0.005$; then $1-\alpha$ is already very close to 1, which makes it less meaningful/useful to impose the constraint of $P ( Y_{test} \in PS(X_{test})) \leq 1-\alpha + \epsilon$.

More importantly, dropping this constraint will allow more flexibility in designing a practically effective CP method. Previously, a CP method was designed and built to be not too conservative and not too relaxed as the gap between the lower and upper constraints in Eq.~(\ref{eq:1}) is often quite small. In critical applications, satisfying the lower end of the constraints is a more essential consideration. Therefore, we aim to impose a stricter criterion when building a CP method to achieve a better guarantee for satisfying Eq.~(\ref{eq:2}) in practice.

The high-level idea of our RR-CP method is to find a reliable region in the prediction set confidence score space for each prediction set size option, where the samples in that region strictly satisfy the correctness constraint in Eq.~(\ref{eq:2}). In test time, generating a prediction set for a given sample is conducted by testing its prediction set confidence scores against the reliable regions thus found, and for the prediction sets in these reliable regions, we choose a prediction set with the smallest size. Below we first describe what the reliable regions are in the prediction set confidence score space and how to estimate the reliable regions using calibration data. Then we 
discuss how to use reliable regions for conformal prediction, and finally provide theoretical analyses of our RR-CP method.
\vspace{-0.3cm}
\subsection{Algorithms and Details}\label{sec:method-p2}
Given a trained classifier $f$ and a calibration set $D_{cali} = (X_{cali}, Y_{cali})$, with $X_{cali} = \{x_1, x_2, \dots, x_n\}$ and $Y_{cali}=\{y_1, y_2, \dots, y_n\}$, we aim to estimate $K$ reliable regions, each corresponding to a set size option for conformal prediction.

\textbf{Preparation.} For each sample $x_i \in X_{cali}$, we apply the classifier $f$ and obtain a vector output $\pi_i$, where each element in $\pi_i[k]$, for $k = 1, 2, \dots, K$, describes the likelihood of the sample $x_i$ belonging to the class $k$. 
We then sort the class probabilities in $\pi_i$ from high to low and obtain $\tau_i = argsort ([\pi_i[1], \pi_i[2], \dots, \pi_i[K]])$, where $\tau_i$ is a permutation of $\{1, ..., K\}$ obtained from the sorting. For a set size option $w$, the prediction set is computed as:
\begin{equation}
PS_i^w = 
\begin{cases}
    \{\tau_i[1]\}                   & \text{if } w=1,\\
    \{\tau_i[1], \dots, \tau_i[w]\} & \text{if } 2\leq w \leq K.
\end{cases}
\label{eq:3}
\end{equation}
\vspace{-0.2cm}
 The prediction set confidence score is computed as:
 \begin{equation}
 C_i^w=\sum_{q=1}^w\pi_i[\tau_i[q]].
 \label{eq:4}
\end{equation} 

For each sample $x_i \in D_{cali}$, we obtain its prediction set $PS_i^w$ and its corresponding confidence $C_i^w$, for $i=1,2, \dots, n$ and $w= 1,2, \dots, K$. 

\textbf{Estimation.} For each prediction set size option, we aim to find a reliable region (for a user-specified error rate $\alpha$) in the prediction set confidence score space. In total, we find $K$ reliable regions. For a given $\alpha$ and a set size option $w$, we collect all the corresponding $PS_i^w$ and $C_i^w$, for $i=1, 2, \dots, n$. We then find $C^w$ with a probability no less than $1-\alpha$, such that \textbf{every sample} that satisfies $C_{i}^w \geq C^w$ has its true label $y_i$ in $PS_i^w$. Note that this criterion is stronger than the following one: With a probability no less than $1-\alpha$, the true label $y_i$ is in $PS_i^w$ for a sample $x_i$ that satisfies $C_{i}^w \geq C^w$. Below we describe how to find/estimate $C^w$ for each set size option $w$. For each value of $w$, given a confidence score $C_{i^*}^w$, we apply Bootstrapping~\cite{efron1994introduction} (using $B$ rounds of sampling with replacement) to check whether with probability $1-\alpha$, $PS_i^w$ contains its true label $y_i$ for \textbf{every} sample $x_i$ that has $C_i^w \geq C_{i^*}^w$. If yes, we record the value $C_{i^*}^w$ in a list $L^w$. We repeat the above process for each value of $C_{i^*}^w$, for $i^*= 1, 2, \dots, n$. We then generate $C^w$ using $C^w = min(L^w)$. Finding the lowest confidence level in this list means to encourage the corresponding reliable region to be large so that in the subsequent inference, the yielded prediction set will be small. We repeat this process to find $C^w$ for each prediction set size option $w$, with $w = 1, 2, \dots, K$. The pseudo-code for this part is given in Alg.~1 and~ Alg.~2. 

\begin{figure}[H]
\centering
\includegraphics[width=0.9\textwidth]{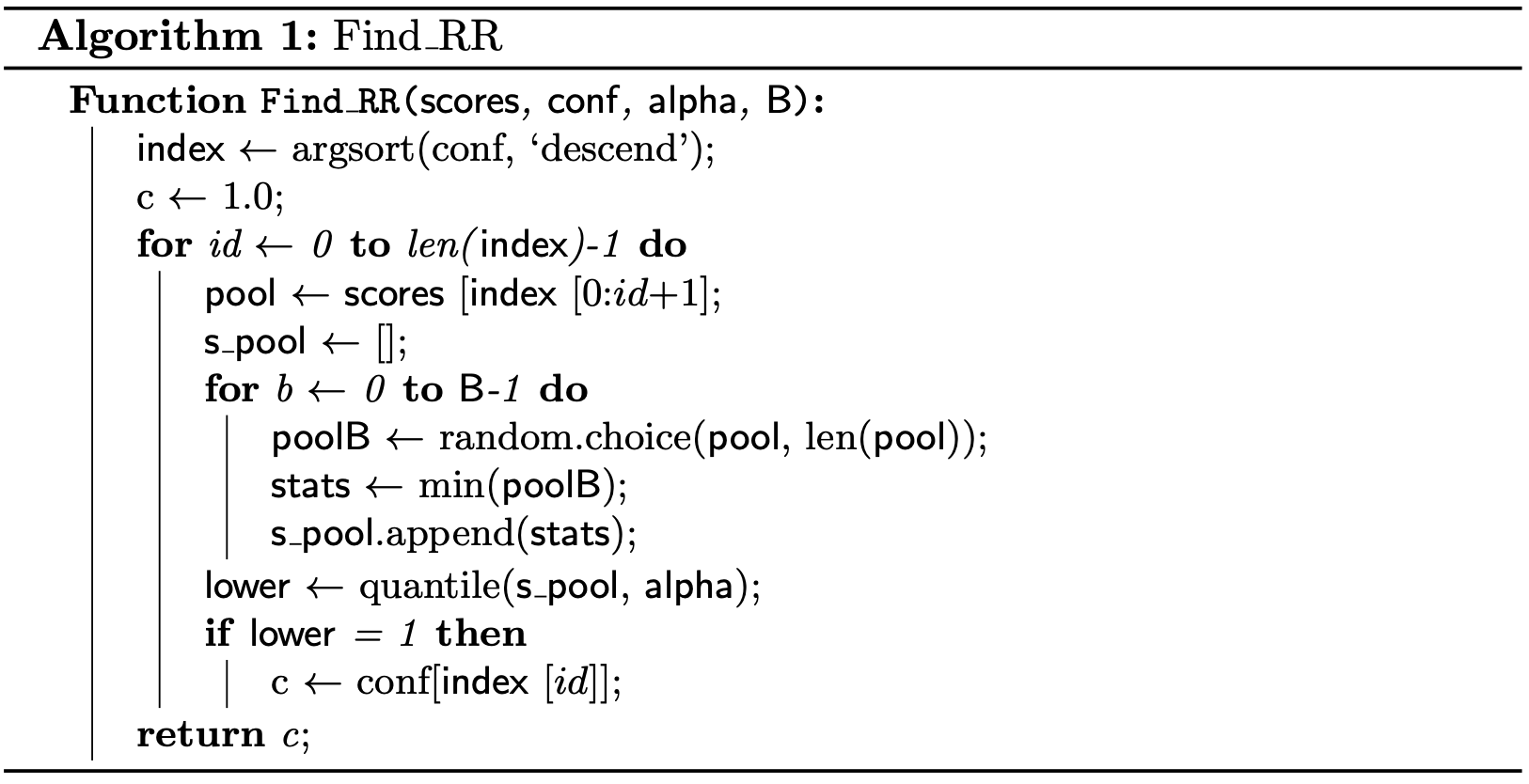}
\label{alg:1}
\end{figure}

\begin{figure}[H]
\centering
\includegraphics[width=0.9\textwidth]{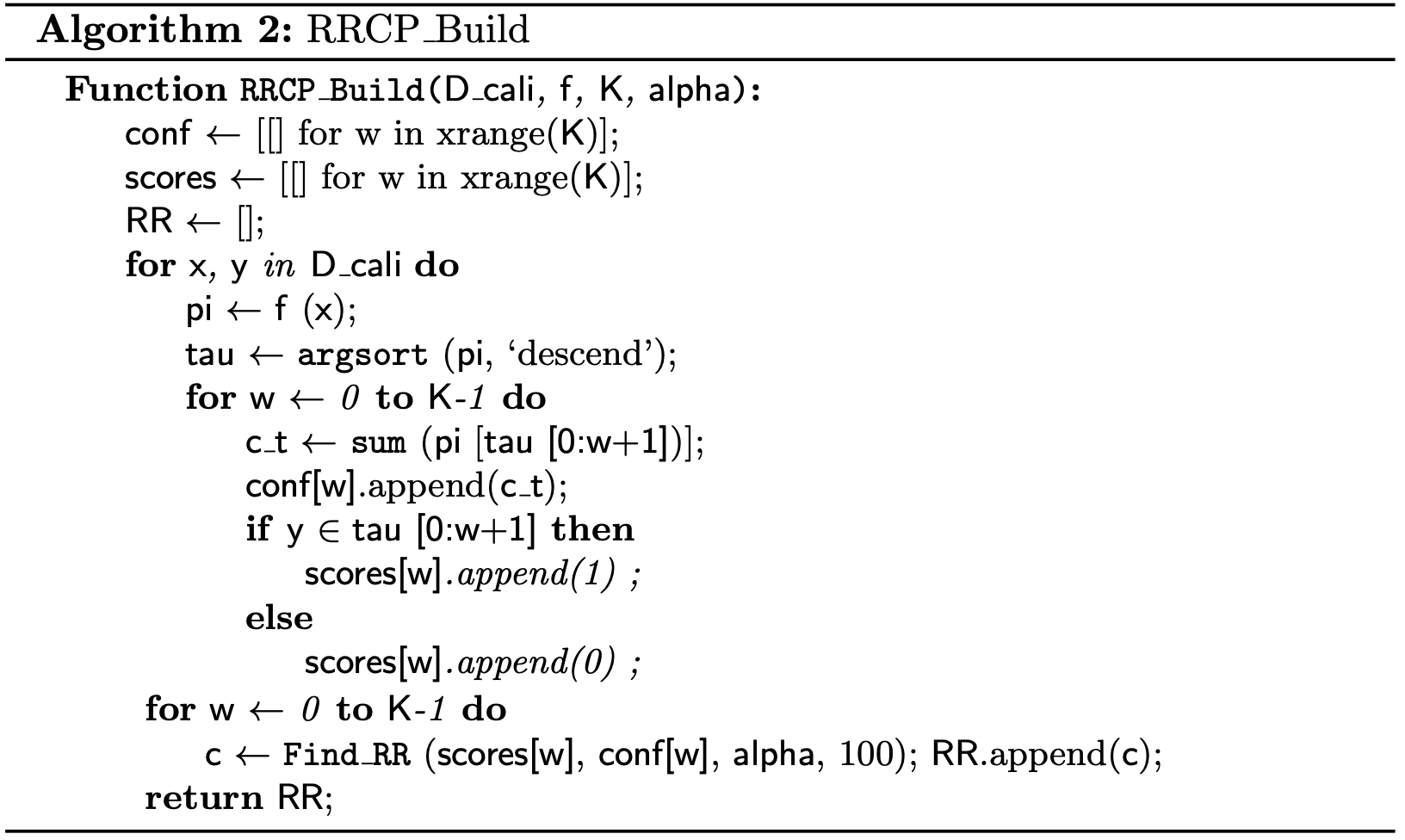}
\label{alg:2}
\end{figure}

\textbf{Inference.} For a new test sample $x_{test}$, we apply $f$ on top of it to obtain $\pi_{test} = f(x_{test})$. We then compute the prediction sets $PS_{test}^w$ and confidence scores $C_{test}^w$ for each set size option (for $w=1,2, \dots, K$) using Eq.~(\ref{eq:4}). With the reliable regions thus found (indicated by $C^w$), we find the smallest possible set size ${w^*}$ which satisfies $C^{w^*} \geq C^w$, and use the obtained ${w^*}$ to report the prediction set $P_{test}^{w^*}$ as the final output. The pseudo-code for using RR-CP to infer the prediction set for a given test sample is given in Alg.~3.

\begin{figure}[H]
\centering
\includegraphics[width=0.9\textwidth]{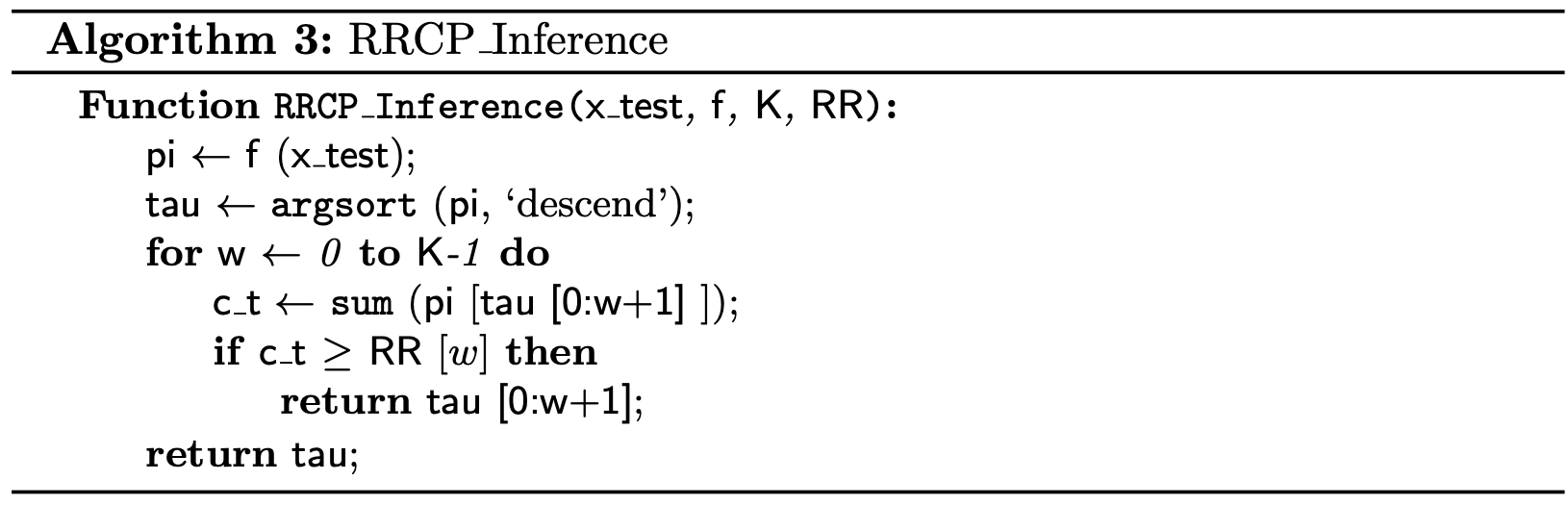}
\label{alg:3}
\end{figure}

\subsection{Theoretical Analyses}\label{sec:method-p3}

\textbf{Proposition:} Suppose the samples in $D_{cali}$ and $D_{test}$ are independent and identically distributed. Then the prediction sets obtained by the RR-CP method (estimated by $D_{cali}$) satisfy \begin{equation}
1- \alpha \leq P ( Y_{test} \in PS(X_{test})) \leq 1 -  \frac{\alpha}{n},
\label{eq:bounds}
\end{equation}
where $n$ is the number of samples in $D_{cali}$. The Bootstrapping loop (line 5 in Listing 1.1) explicitly checks the correctness of the samples in the candidate regions for determining the reliable regions, where the correctness check considers all the prediction sets (PSs) in a given candidate confidence score region. If the $\alpha$ quantile (line 9 in Listing 1.1) obtained from the statistics of the Bootstrapping loop is equal to 1 (line 10 in Listing 1.1), then it means that with a $1-\alpha$ chance, all the PSs in this region are correct (i.e., containing true labels). As a result, the $1- \alpha \leq P (Y_{test} \in PS(X_{test}))$ part is enforced in the worst case scenario: for the $\alpha$ chance when not all the PSs in the region are correct, all the PSs are incorrect. In this worst case, $P (Y_{test} \in PS(X_{test})) = 1- \alpha$. In the best case scenario, for the $\alpha$ chance when not all the PSs in the region are correct, only 1 PS 
is incorrect. Hence, the best case gives the part $P ( Y_{test} \in PS(X_{test})) \leq 1 -  \frac{\alpha}{n}$. Note that with more samples in $D_{cali}$ ($n$ getting larger), the upper-bound of $P ( Y_{test} \in PS(X_{test}))$ is approaching 1. Although the gap between the lower-bound and upper-bound in Eq.~(\ref{eq:bounds}) is not as tight as the bounds from the previous CP methods (e.g., Eq.~(\ref{eq:1}) from~\cite{vovk1999machine}), RR-CP still gives a pair of bounds that is practically tight enough for small values of $\alpha$.\footnote{The user-specified error rate $\alpha$ should be very small for medical application cases.} More importantly, RR-CP considers the worst-case scenario for the reliable regions when establishing the lower-bound of the coverage (with $1-\alpha$), making it more viable to achieve such a bound.

\subsection{Limitations}
The proposed RR-CP method is designed to achieve an extremely low actual error rate in model deployment for critical applications, such as medical diagnosis. However, the limitation of RR-CP becomes evident in situations where the user-set error rate is high, such as $5\%$ or $10\%$. This issue arises because the gap between the lower and upper bounds in Eq.~\ref{eq:bounds} increases as $\alpha$ grows larger. Consequently, when dealing with a large user-specified error rate ($\alpha$), RR-CP might yield an actual error rate that is significantly smaller than the user-specified rate. This particular aspect of RR-CP could be considered a drawback from the conventional perspective of CP methods.

\section{Experiments}

\begin{table*}[t]
\footnotesize
\begin{center}
\caption{Performance comparisons of our RR-CP and known CP methods. The user-specified error rate $\alpha$ is set to be $0.5\%$. All the methods are evaluated using the test sets of the datasets. The actual error rates that are equal to or smaller than $0.5\%$ are marked in {\bf bold}. AER: actual error rate ($\%$). PSS: prediction set size. }
\begin{tabular}{|c | c | c | c| c | c | c| c| c| c| c|} 
 \hline
  \multirow{2}{*}{Method}  & \multicolumn{2}{|c|}{OrganA} & \multicolumn{2}{|c|}{Blood}& \multicolumn{2}{|c|}{OrganC} & \multicolumn{2}{|c|}{OrganS} & \multicolumn{2}{|c|}{Derma}\\
 \cline{2-11}
  &  AER &  PSS & AER &  PSS  & AER &  PSS & AER &  PSS & AER &  PSS \\
\hline
CP~\cite{vovk1999machine} & 0.78 & 1.5 & 0.85 & 1.2 & 2.09 & 1.8 & 2.25 & 2.7 & 0.85 & 3.2\\
\hline
RAPS~\cite{angelopoulosuncertainty} & 0.79 & 1.7 &  0.59 & 1.3& 2.19 & 1.8 & 1.60& 3.2 & \textbf{0.16} & 4.27 \\
\hline
ERAPS~\cite{xuERPAS2022} & 0.96 & 6.2 &  0.74 & 2.0& 0.82 & 8.3 & 0.88& 7.1 &0.55 & 5.2\\
\hline
RR-CP (ours) & \textbf{0.42} & 2.4 & \textbf{0.39} & 1.6 & {1.38} & {2.1} & \textbf{0.49} & 4.8 & \textbf{0.15} & 4.28 \\
\hline

\end{tabular}
\label{table:result1}
\end{center}
\end{table*}

\subsection{Datasets and Setups}
 Five public medical image classification datasets, OrganAMNIST (11 classes)~\cite{bilic2023liver}, OrganCMNIST (11 classes)~\cite{bilic2023liver}, OrganSMNIST (11 classes)~\cite{bilic2023liver}, BloodMNIST (8 classes)~\cite{acevedo2020dataset}, and DermaMNIST (7 classes)~\cite{tschandl2018ham10000} (all obtained from the MedMNIST benchmark~\cite{medmnistv2}) are utilized for the experiments. The numbers of samples in the train/val/test sets are 34581/6491/17778 for OrganAMNIST, 13000/2392/8268 for OrganCMNIST, 13940/2452/8829 for OrganSMNIST, 11959/1712/3421 for BloodMNIST, and 7007/1003/2005 for DermaMNIST. For a particular dataset, its training set is used to train a classifier. Here, we use ResNet18 \cite{he2016deep} as the classifier for all the experiments. Model weights are obtained from the official release\footnote{https://github.com/MedMNIST/experiments}. The validation set is used as the set $D_{cali}$ for estimating the parameters in a CP method, and the test set is used to evaluate the CP method for the actual error rate and its corresponding prediction set size. We compare our RR-CP with the classic CP method~\cite{vovk1999machine}, the recently developed Regularized Adaptive Prediction Set (RAPS)~\cite{angelopoulosuncertainty}, and Ensemble-RAPS~\cite{xuERPAS2022}. As discussed above, $\alpha$ (the user-specified/chosen error rate) should be set as a small value for medical use cases. We set $\alpha$ = 0.005 for all the methods utilized in the experiments.  
\vspace{-0.05in}
\subsection{Results}

\textbf{On Achieving the User-specified Error Rate.} In Table~\ref{table:result1}, we show that our RR-CP achieves the desired error rate (0.005) on four (out of five) datasets, and the competing methods failed to achieve the desired error rate for most of the experiments. Particularly, on OrganAMNIST, RR-CP achieves almost half of the error rates compared to the competing methods. On BloodMNIST, RR-CP gives $54\%$, $47\%$, and $34\%$ reductions in the error rate compared to the classic CP, ERAPS, and RAPS, respectively. On OrganCMNIST, all the known CP methods failed to achieve the user-chosen error rate. On OrganSMNIST, RR-CP achieves the desired error rate, and its actual error rate is also significantly lower than those of the competing methods. On DermaMNIST, the classic CP and ERAPS failed to meet the user-specified error rate, while both RR-CP and RAPS satisfy the specified error rate on this dataset.

\textbf{On Prediction Set Size.} Due to the stronger guarantee of satisfying the user-specified error rate, RR-CP is observed to give a relatively large prediction set (see the ``PSS'' columns in Table~\ref{table:result1}). As discussed and emphasized in the previous sections, smaller-sized prediction sets are meaningful only when the desired error rate is honored and achieved in deployment. This is especially true and critical for medical and diagnostic use cases, where trading accuracy for smaller prediction set sizes is often unacceptable. The larger-sized prediction sets from RR-CP are needed for the strong guarantee of satisfying the user-specified low error rate (and RR-CP achieves it in most of the experiments). It is worth noting that when RR-CP and RAPS yield a similar level of actual error rate (see the ``Derma'' column in Table~\ref{table:result1}), their prediction set sizes are almost the same. This indicates that the sizes of prediction sets generated by RR-CP are reasonably small with respect to the actual error rates it attains. In addition, on OrganA/C/S, we observe that ERAPS yields prediction sets with sizes bigger than half of the largest possible set size. Too relaxed prediction sets, as in the ERAPS case, can make CP less effective for computer-aided diagnosis.

\textbf{Acknowledgement.} This work was supported in part by National Natural Science Foundation of China (62201263) and Natural Science Foundation of Jiangsu Province (BK20220949). S.W. is supported by Shanghai Sailing Programs of Shanghai Municipal Science and Technology Committee (22YF1409300).

\section{Conclusions}
In this paper, we identified the known CP methods often failed to reach a desired low error rate for medical image classification. We developed a novel, statistically sound, and effective method (RR-CP) for addressing this problem. RR-CP is designed to impose a strong correctness guarantee. Theoretical analyses showed that in the worst case scenario, RR-CP yields an error rate that is the same as the user-specified error rate. Empirical results further validated that, on real data on which the i.i.d. assumption is not exactly held, RR-CP achieved the desired low error rate significantly more often than the known CP methods.

\bibliographystyle{plain}
\bibliography{ref}
\end{document}